\documentclass[conference]{IEEEtran}
\IEEEoverridecommandlockouts
\usepackage{cite}
\usepackage{array}
\usepackage{booktabs}
\usepackage{amsmath,amssymb,amsfonts}
\usepackage{algorithmic}
\usepackage{graphicx}
\usepackage{textcomp}
\usepackage{xcolor}
\def\BibTeX{{\rm B\kern-.05em{\sc i\kern-.025em b}\kern-.08em
    T\kern-.1667em\lower.7ex\hbox{E}\kern-.125emX}}

\usepackage{fancyhdr}
\begin{document}

\title{Advancing Pain Recognition through Statistical Correlation-Driven Multimodal Fusion*\\
\thanks{\textbf{This research is supported by Holistic AI and University College London.} Corresponding Author: Zekun Wu}}

\author{\IEEEauthorblockN{Xingrui Gu}
\IEEEauthorblockA{\textit{Electrical Engineering and Computer Science} \\
\textit{University of California, Berkeley}\\
Berkeley, CA \\
xingrui\_gu@berkeley.edu}
\and
\IEEEauthorblockN{Zhixuan Wang}
\IEEEauthorblockA{\textit{Department of Computer Science} \\
\textit{University College London}\\
London, UK \\
ggzhwzx@gmail.com}
\and
\IEEEauthorblockN{Irisa Jin}
\IEEEauthorblockA{\textit{Halıcıoğlu Data Science Institute} \\
\textit{University of California, San Diego}\\
San Diego, United States \\
irisajin23@gmail.com}
\and
\IEEEauthorblockN{Zekun Wu}
\IEEEauthorblockA{\textit{Holistic AI}\\
\textit{University College London}\\
\textit{London, UK}\\
zekun.wu@holisticai.com}}

\maketitle
\thispagestyle{fancy}

\begin{abstract}
This research presents a novel multimodal data fusion methodology for pain behavior recognition, integrating statistical correlation analysis with human-centered insights. Our approach introduces two key innovations: 1) integrating data-driven statistical relevance weights into the fusion strategy to effectively utilize complementary information from heterogeneous modalities, and 2) incorporating human-centric movement characteristics into multimodal representation learning for detailed modeling of pain behaviors. Validated across various deep learning architectures, our method demonstrates superior performance and broad applicability. We propose a customizable framework that aligns each modality with a suitable classifier based on statistical significance, advancing personalized and effective multimodal fusion. Furthermore, our methodology provides explainable analysis of multimodal data, contributing to interpretable and explainable AI in healthcare. By highlighting the importance of data diversity and modality-specific representations, we enhance traditional fusion techniques and set new standards for recognizing complex pain behaviors. Our findings have significant implications for promoting patient-centered healthcare interventions and supporting explainable clinical decision-making.

\end{abstract}


\section{Introduction}

\par
\begin{figure}[h]
    \centering
    \includegraphics[width=0.5\textwidth]{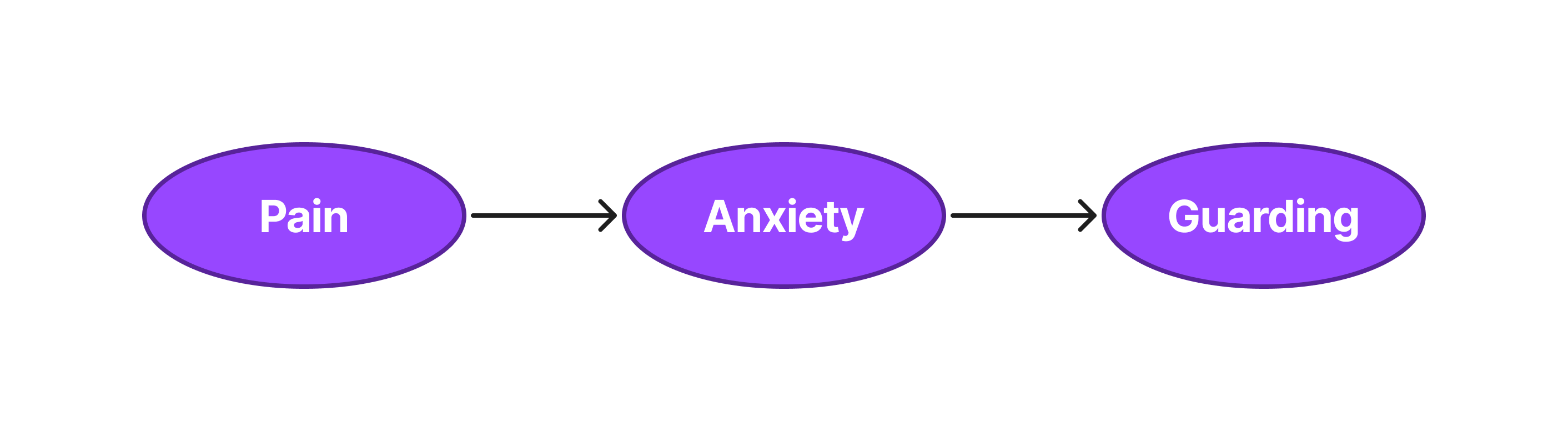}
    \caption{Relationship between pain, emotion and protective behaviour\cite{olugbade2019relationship}}
    \label{fig:painemotion}
\end{figure}

 Affective Computing, an interdisciplinary field within Human-Computer Interaction, holds immense potential for enhancing human-computer interfaces and health surveillance by recognizing and responding to human emotions\cite{picard2000affective}. Its application in understanding human pain behavior is particularly promising, as it acknowledges pain as a complex emotional state rather than merely a physical condition \cite{abdullah2021multimodal}\cite{wade1990emotional}. As illustrated in Figure \ref{fig:painemotion}, pain is closely intertwined with anxiety, which can lead to protective behaviors such as guarding\cite{olugbade2019relationship}. However, accurate pain recognition remains challenging due to the multi-dimensionality and subjectivity of pain experiences, which stem from the intricate interplay of physiological, psychological, and social factors\cite{wang2022systematic}\cite{barrett2017emotions}\cite{khera2021cognition}.

\par Multimodal data fusion has emerged as a powerful approach in human-centered computing to address these challenges by integrating information from diverse sources such as physiological signals, behavioral cues, and self-reports\cite{arrieta2020explainable}. Nevertheless, existing multimodal pain recognition methods often struggle with data heterogeneity, alignment issues, and limited interpretability\cite{salekin2022attentional}. Moreover, conventional machine learning and deep learning techniques often fail to capture the complexities of pain experiences and lack the adaptability to personalize pain recognition\cite{gao2020survey}.

\par To overcome these limitations, this research introduces a novel approach to multimodal pain recognition that synergistically integrates statistical methods with a human-centric perspective. Our key innovations lie in: 1) employing statistical inference and hypothesis testing to explore the relationships between different modalities and the target variable (pain states), identifying the most informative features for pain recognition; and 2) incorporating human-centered insights into the representation and modeling of pain experiences to ensure the interpretability and ethical alignment of our models. By dynamically adjusting the contributions of different modalities through adaptive weighting, our data-driven, personalized approach enhances the precision, efficiency, and adaptability of pain recognition. Ultimately, this research contributes to the advancement of pain recognition, affective computing, AI-assisted healthcare, and the development of empathetic and socially responsible AI systems, paving the way for more effective and ethically grounded pain management strategies. 

\section{Related Work}

\par Previous studies have suggested that traditional machine learning and deep learning approaches can serve as alternatives to manual pain scoring. For instance, using a conventional neural network based on Google’s InceptionV3 model, researchers predicted binary labels of “pain” and “no pain” from mouse facial images \cite{tuttle2018deep}. Additionally, Surface Electromyography (SEMG) recordings have proven valuable in assessing chronic low back pain \cite{ambroz2000chronic}. Building on this foundational research, a new perspective in pain recognition emphasizes the integration of multimodal data to leverage the strengths of various sources, thereby enhancing the accuracy and precision of predicting pain-related behaviors. However, it is crucial first to establish and evaluate pain as a human emotion. Previous research has identified overlapping brain regions within the Central Nervous System (CNS) that are involved in both pain and other emotions, including the amygdala, thalamus, and Anterior Cingulate Cortex (ACC) \cite{gilam2020relationship}.


\par Building upon this foundation of research, there emerges a novel perspective within affective computing and human centered computing, wherein the integration of multimodal data can use the strengths of various sources of data to enhance the accuracy and precision of predicting pain-related behavior. The recognition of emotion and pain requires a nuanced understanding of complex constructs, synthesizing from a diverse array of modal features, such as facial expressions, vocal tones, postures, physiological signals, sensory experiences, emotional states, and cognitive evaluations \cite{barrett2017emotions}\cite{khera2021cognition}. 

\par This complexity necessitates an integrated, multimodal analytical approach that goes beyond singular data sources. Consequently, state-of-the-art deep learning algorithms, including CNN-LSTMs, have been leveraged to significantly improve the accuracy and speed of pain recognition \cite{haque2018deep}\cite{werner2014automatic}. However, the challenge of effectively integrating and processing this heterogeneous data persists, and overcoming this challenge is essential for advancing multimodal data fusion and maximizing the efficacy of analytical methods for pain recognition \cite{gao2020survey}.

\par Reflecting on this necessity, historical research reinforces the value of integrating multiple data types, such as body movements and muscle activity, to enhance pain recognition accuracy \cite{olugbade2014bi}. These findings highlight the crucial interconnectivity of different modalities, underpinning their essential role in pain recognition and supporting the advancement of multimodal methodologies in this study.

\par Subsequent explorations have further elucidated this concept, with studies illustrating the efficacy of combining muscle and motion signals for the accurate identification of protective behaviors—a key aspect of pain recognition. Notably, central (model-level) fusion approaches have been shown to outperform both feature and decision-level fusion methods in the context of Protective Behavior Detection (PBD), showcasing their superior capability in harnessing the potential of multimodal data\cite{wang2021chronic}\cite{cen2022exploring}. This advancement signals a critical insight into the differential impact of various fusion layers on the process of pain recognition, highlighting the imperative of selecting and implementing the most effective fusion strategies to achieve a more integrative and holistic understanding of pain behaviors through the synthesis of multimodal information.

\par Innovative model architectures have been developed to enhance the accuracy and applicability of pain recognition studies. The P-STEMR framework is a notable example, utilizing human activity recognition datasets to classify pain levels, showcasing the utility of supervised learning in situations with limited labeled data sources \cite{olugbade2023movement}. Additionally, the introduction of an advanced hierarchical HAR-PBD architecture represents significant progress in real-time pain recognition. This architecture combines Human Activity Recognition (HAR) with Protective Behavior Detection (PBD) using graph convolution and Long Short-Term Memory (LSTM) networks. Moreover, this model addresses the common challenge of data classification imbalance by implementing the Class-Balanced Focal Classification Cross-Entropy (CFCC) loss function \cite{wang2021leveraging}, thereby enhancing the reliability and effectiveness of pain monitoring systems.


\par The challenges in multimodal data fusion involve integrating heterogeneous data from diverse sources for accurate pain state recognition, where increased dimensionality can reduce model interpretability. While previous research has employed machine learning and deep learning for automated pain recognition, these methods often struggle with complex, multi-source pain behaviors. Our novel approach emphasizes a multimodal data fusion strategy grounded in statistical relevance, complemented by a human-centric perspective to enhance model effectiveness. This methodology not only accounts for data diversity but also transforms the relationships between different modalities and outcomes into weighted contributions for model decision-making, resulting in more precise and comprehensive detection of pain behaviors.

\section{Methodology}
\par In this study, our objective is to investigate and substantiate the efficacy and relevance of statistical approaches in amalgamating multimodal data, with a particular focus on domains centered around human computation, such as the analysis of pain. The endeavor is to amalgamate a variety of statistical instruments to refine the integration process of multimodal data, thereby augmenting the precision and efficiency in the analysis of pain behaviors. We postulate that employing this strategy will facilitate a more nuanced comprehension of the intricacies and multifaceted nature of pain, consequently enabling the provision of solutions for pain management that are both more targeted and individualized.

\par The central focus of this research is on developing methodologies for the effective fusion and processing of multimodal data from diverse sources and types. This involves utilizing statistical techniques to integrate modal features, conducting hypothesis tests on target variables, and performing correlation analyses. Additionally, the study aims to evaluate the significance of different modalities and determine dynamic weight distribution based on statistical insights. It also seeks to optimize the combination of features to improve the accuracy and effectiveness of models in predicting pain.

\subsection{Dataset Introduction}
\begin{figure}[h]
    \centering
    \includegraphics[width=0.4\textwidth]{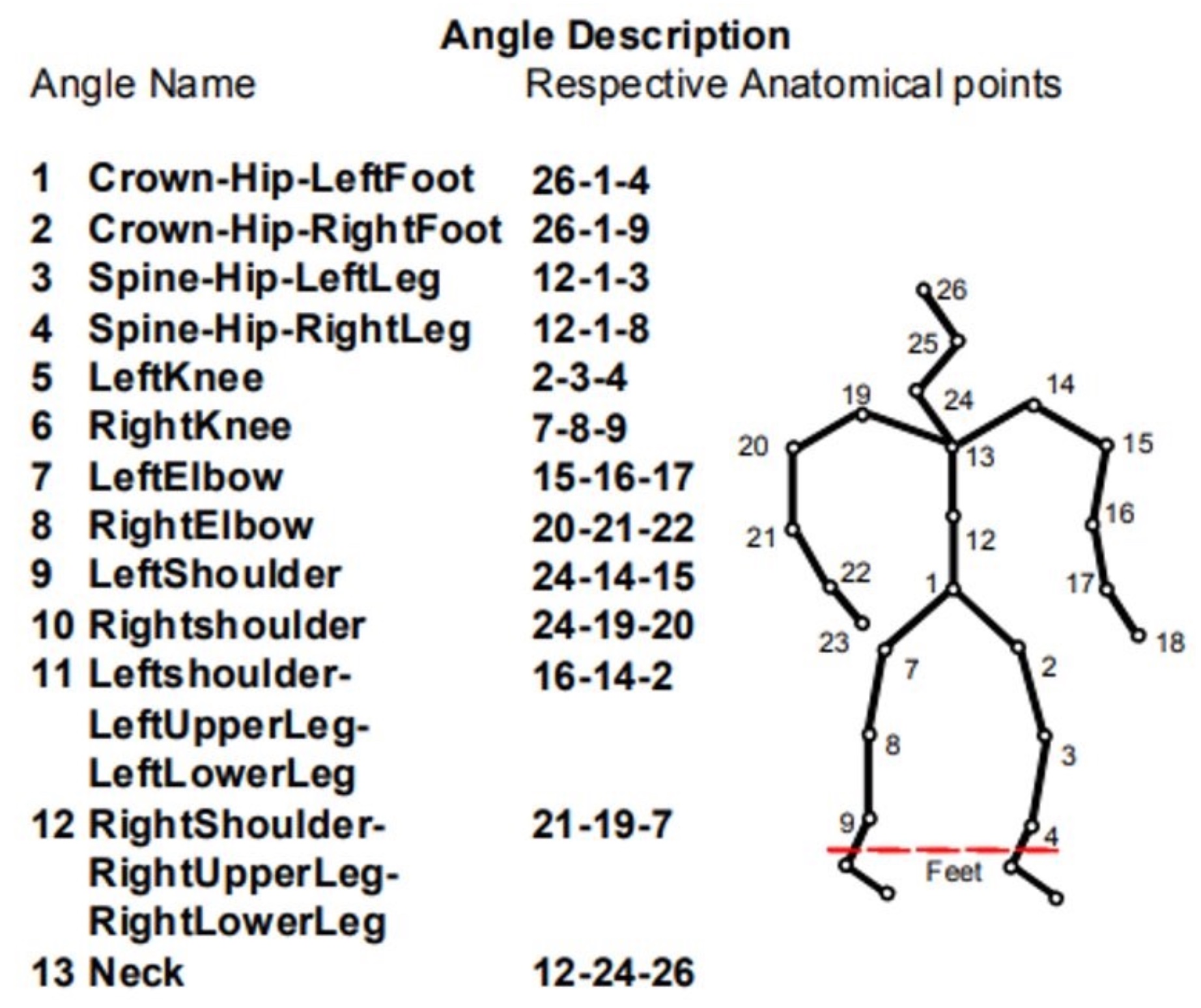}
    \caption{The arrangement of the 22 body joints\cite{10.1145/3463508}}
    \label{fig:relation}
\end{figure}

In this research, we utilized the EmoPain dataset, a key resource for exploring the relationship between body movements and pain intensity levels \cite{aung2015automatic}. The dataset is divided into training and validation sets, with data from 10 chronic pain sufferers and 6 healthy controls in the training set, and 4 chronic pain individuals and 3 healthy controls in the validation set. As detailed in Table \ref{tab:data-file-measures} and Figure \ref{fig:relation}, the dataset includes X, Y, and Z coordinates of body joints, categorized in columns 1-22, 23-44, and 45-66, respectively. The core of our analysis focuses on vector 73, which measures protective behavior, distinguishing non-protective actions (coded as 0) from protective behaviors (coded as 1). This complex interplay between protective behaviors and pain, mediated by emotional states, positions our study at the intersection of behavioral analysis, pain recognition, and emotional computation.


\begin{table}[h!]
\centering
\begin{tabular}{@{}lp{7.5cm}@{}}
\toprule
\textbf{Columns} & \textbf{Description} \\ \midrule
1-22 & X coordinates of 22 body joints. \\
23-44 & Y coordinates of 22 body joints. \\
45-66 & Z coordinates of 22 body joints. \\
67-70 & Surface electromyography data from the lumbar and upper trapezius muscles. \\
73 & Protective behaviour label (0 for not protective, 1 for protective). \\ \bottomrule
\end{tabular}
\label{tab:data-file-measures}
\end{table}

\begin{table}[h]
\centering
\caption{Comparison of Statistical Methods}
\label{tab:Comparison of Statistical Methods}
\begin{tabular}{@{}p{1cm}p{3.5cm}p{3.5cm}@{}}
\toprule
\textbf{Method} & \textbf{Usage} & \textbf{Data Distribution} \\ \midrule
ANOVA & Independent groups & Normal, Equal variances \\
\midrule
Pearson & Linear relation & Normal \\
\midrule
Spearman & Monotonic relation & Any, including non-normal \\
\midrule
Kendall & Ordered pairs, small
sample size or with tied ranks & Any, including non-normal \\
\bottomrule
\end{tabular}
\end{table}

\begin{figure}[h]
    \centering
    \includegraphics[width=0.4\textwidth]{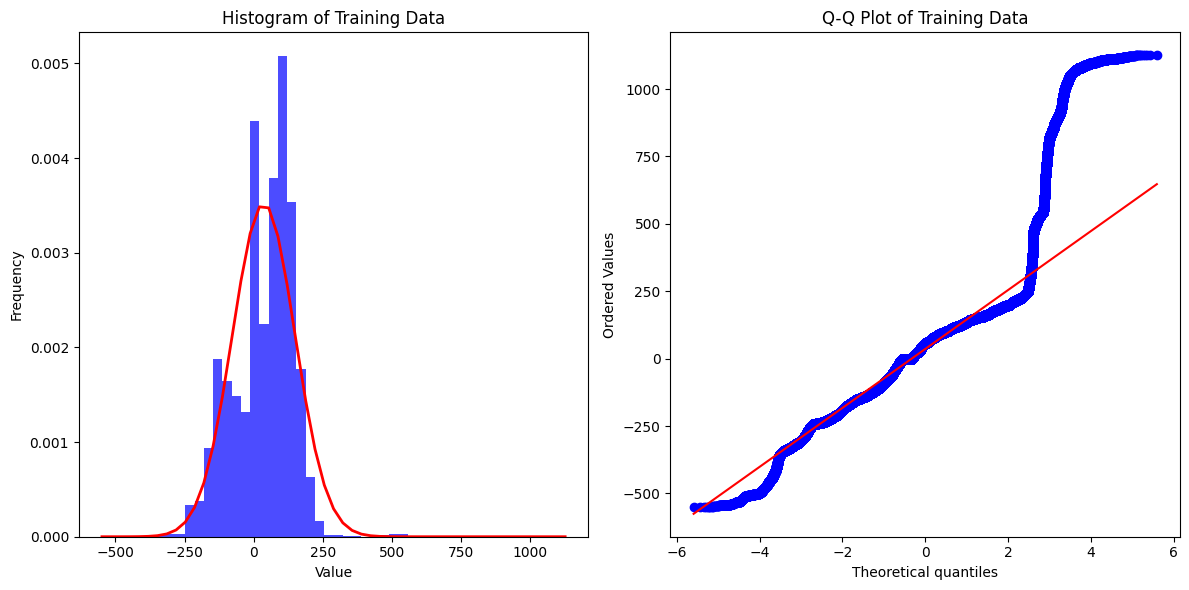} 
    \hspace{1cm} 
    \includegraphics[width=0.4\textwidth]{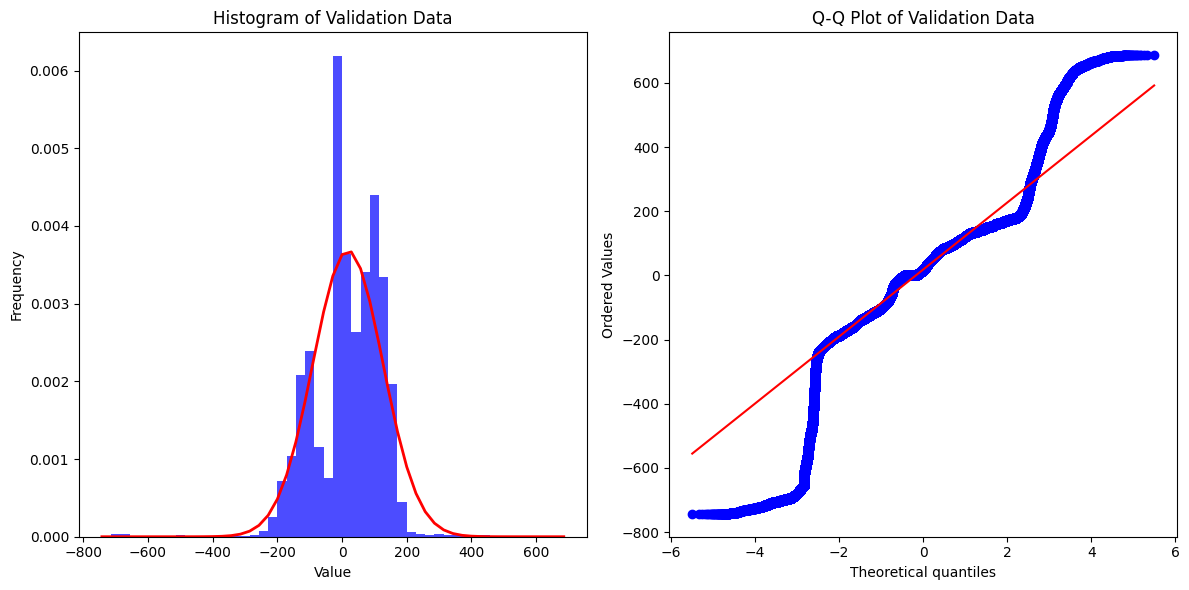} 
    \caption{Left: Train Data Distribution. Right: Valid Data Distribution.}
    \label{fig:normal}
\end{figure}

    
    

\subsection{Dataset Analysis}
\par The histograms and Q-Q plots in Figure \ref{fig:normal}  for the training and validation data provide crucial insights into the normality of the dataset's distribution. From the histograms, both datasets display a pronounced peak, but it doesn't align with the theoretical normal distribution curve, suggesting a discrepancy from normality. This is further evidenced by the asymmetry and apparent deviations from the central peak.

\par The Q-Q plots reinforce these findings. Ideally, data points should closely follow the red line if they were normally distributed. However, in both training and validation datasets, significant deviations occur, particularly in the tails of the distribution. These deviations manifest as pronounced curves away from the expected line, indicating heavier tails than those of a normal distribution and suggesting a skew in the data.

\par Given the distribution characteristics of our dataset, as outlined in Table \ref{tab:Comparison of Statistical Methods}, we identify the limitations and suitability of various statistical methods for our analysis. The assumption of normality renders \textbf{ANOVA} and \textbf{Pearson correlation} ineffective for our purposes, while \textbf{Kendall's rank correlation}, despite its robustness, is impractical due to computational demands and is more suited for smaller datasets. Consequently, we opt for \textbf{Spearman's rank correlation} as the most fitting choice, owing to its efficiency with large datasets and applicability to non-linear relationships, ensuring a more accurate and relevant analysis of our multimodal data. This decision strategically aligns with our analytical needs, highlighting Spearman’s correlation as the optimal tool for exploring the complex relationships within our heterogeneous dataset.

\subsection{Experiment Design}
\par The experimental design rigorously explores the influence of diverse data fusion strategies on the effectiveness of protective behavior recognition models. It delves into how decision-level fusion, enhanced by statistical analysis of data heterogeneity and human-centered modalities, affects model performance in the spectrum of protective behavior identification.

\subsubsection{Decision-Level Fusion with Singular Modality (Benchmark Model)}
\begin{figure}[h]
    \centering
    \includegraphics[width=0.45\textwidth]{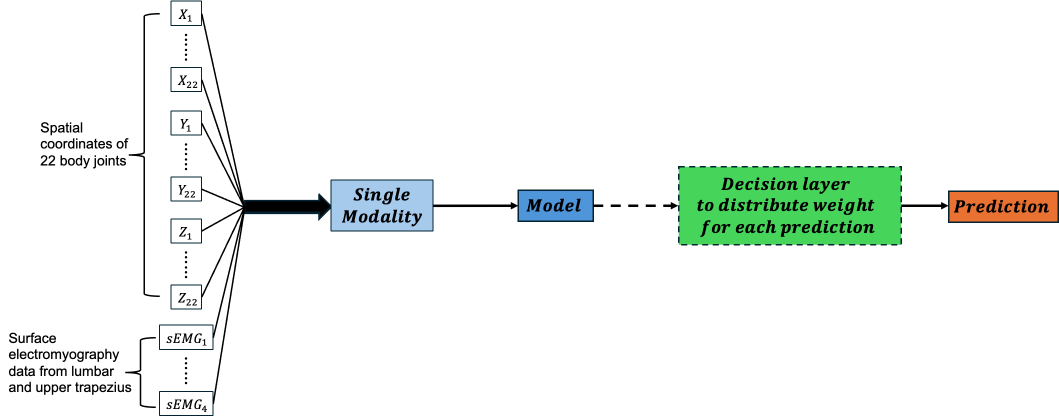}
    \caption{Singular Modality (Benchmark Model)}
    \label{fig:singular}
\end{figure}

The foundational experiment initiates with all 70 features amalgamated as a singular modality, subjected to a single Classification model. This setup, while ostensibly a feature-level fusion, is underscored as a decision-level fusion where the entire feature set is implicitly accorded a 100\% weightage. This model establishes the baseline for performance metrics against which the outcomes of subsequent fusion techniques are compared (see Figure \ref{fig:singular}).

\subsubsection{Bifurcated Modality Approach}
\begin{figure}[h]
    \centering
    \includegraphics[width=0.45\textwidth]{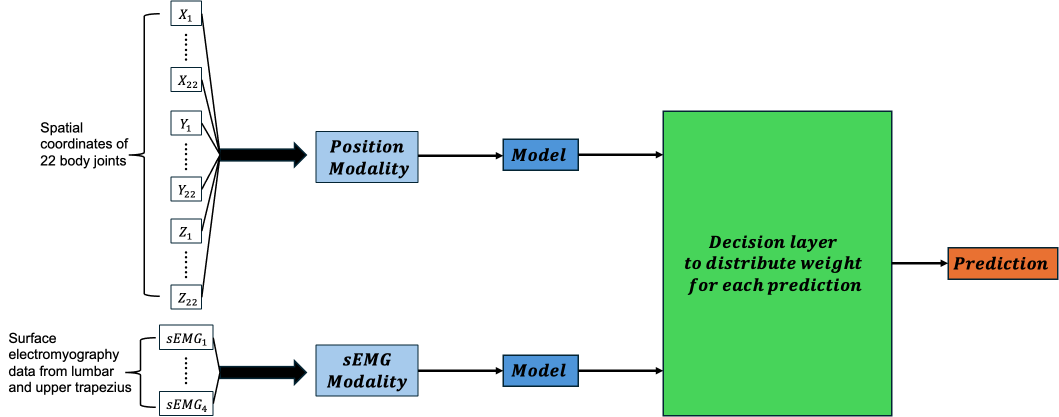}
    \caption{Bifurcated Modality Approach}
    \label{fig:bifurcated}
\end{figure}

Progressing to acknowledge the heterogeneity inherent in the data, the experiment bifurcates the features into two distinct modalities: the XYZ coordinates and sEMG signals. Each modality is then independently processed through an identical Classification model framework. Post-training, a weighted voting mechanism—rooted in decision-level fusion—is employed to amalgamate the predictive insights from each modality. The weights are meticulously derived from Spearman rank correlation coefficients, mirroring the relative significance of each modality's contribution to pain level prediction. This phase aims to unravel whether segmenting features based on their heterogeneity and integrating them through a weighted decision-making process can elevate the model's performance beyond the benchmark (see Figure \ref{fig:bifurcated}).

\subsubsection{Quadrifurcated Modality Approach (Incorporating Human Factors)}
\begin{figure}[h]
    \centering
    \includegraphics[width=0.45\textwidth]{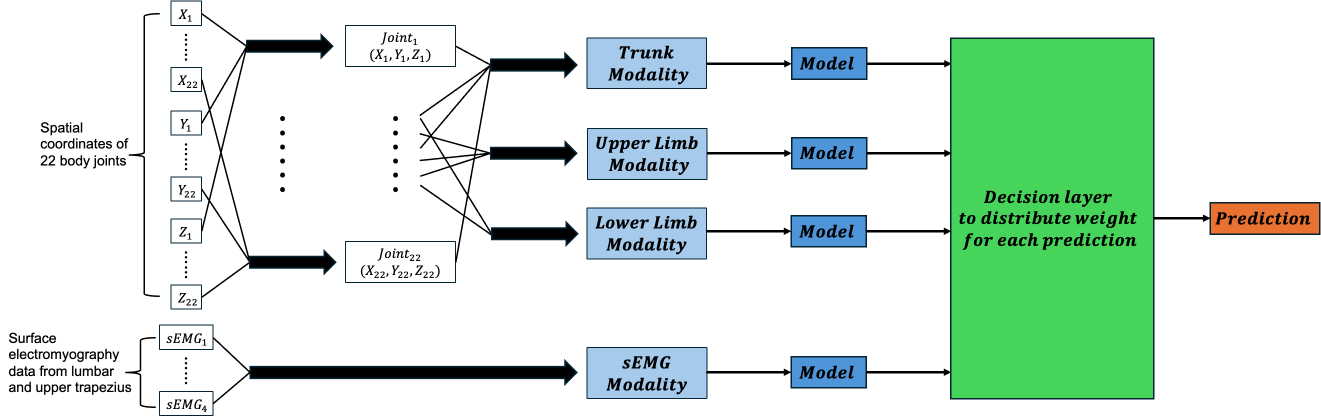}
    \caption{Quadrifurcated Modality Approach (Incorporating Human Factors)}
    \label{fig:quadrifurcated}
\end{figure}

Further dissecting the data through a lens of human factors, this segment of the experiment categorizes the XYZ coordinates into three additional modalities, each representing a distinct body segment (upper limbs, lower limbs, and trunk), alongside the sEMG modality, culminating in a four-modality framework. Each modality, processed through the same Classification model, undergoes an analogous weighted voting system post-training, with weights again anchored in Spearman rank correlation coefficients(See Figure \ref{fig:quadrifurcated}).

\subsubsection{Average Weighted Voting as a Comparative Benchmark}
To enhance the comparison within our study, we have implemented an average weighted voting mechanism across the four modalities, which presumes each modality contributes equally, without considering their statistical correlation to the protective behaviour states being analysed. This baseline method is essential for our analysis as it starkly contrasts with the statistically weighted voting approach, thereby underscoring the advantages of utilizing statistical correlations for modality weighting. This comparative analysis not only highlights the efficacy of statistical correlations but also emphasizes the significance of adopting a human-centered perspective in modality segmentation, showcasing its potential to yield more nuanced and accurate emotion recognition results.

\subsubsection{Comparative Analysis \& Model Performance Evaluation}
By comparing the performance of the singular modality (benchmark model) with both the bifurcated and quadrifurcated modality approaches, as well as the average weighted voting benchmark, the experiment aspires to illuminate whether a granular, human-centered feature segmentation supplemented by statistically weighted decision-level fusion markedly optimizes model accuracy in pain recognition. The transition from a singular, homogenously weighted modality to a nuanced, statistically or evenly weighted integration of multiple modalities endeavors to elucidate the symbiotic relationship between data-driven and human-centered segmentation, and their collective prowess in enhancing pain recognition models. This comprehensive exploration aims to crystallize the efficacy of each strategic approach and underscore their cumulative contribution towards refining model performance in the nuanced domain of pain recognition.s

\section{Results and Evaluation}
\begin{figure}[h]
    \centering
    \includegraphics[width=0.45\textwidth]{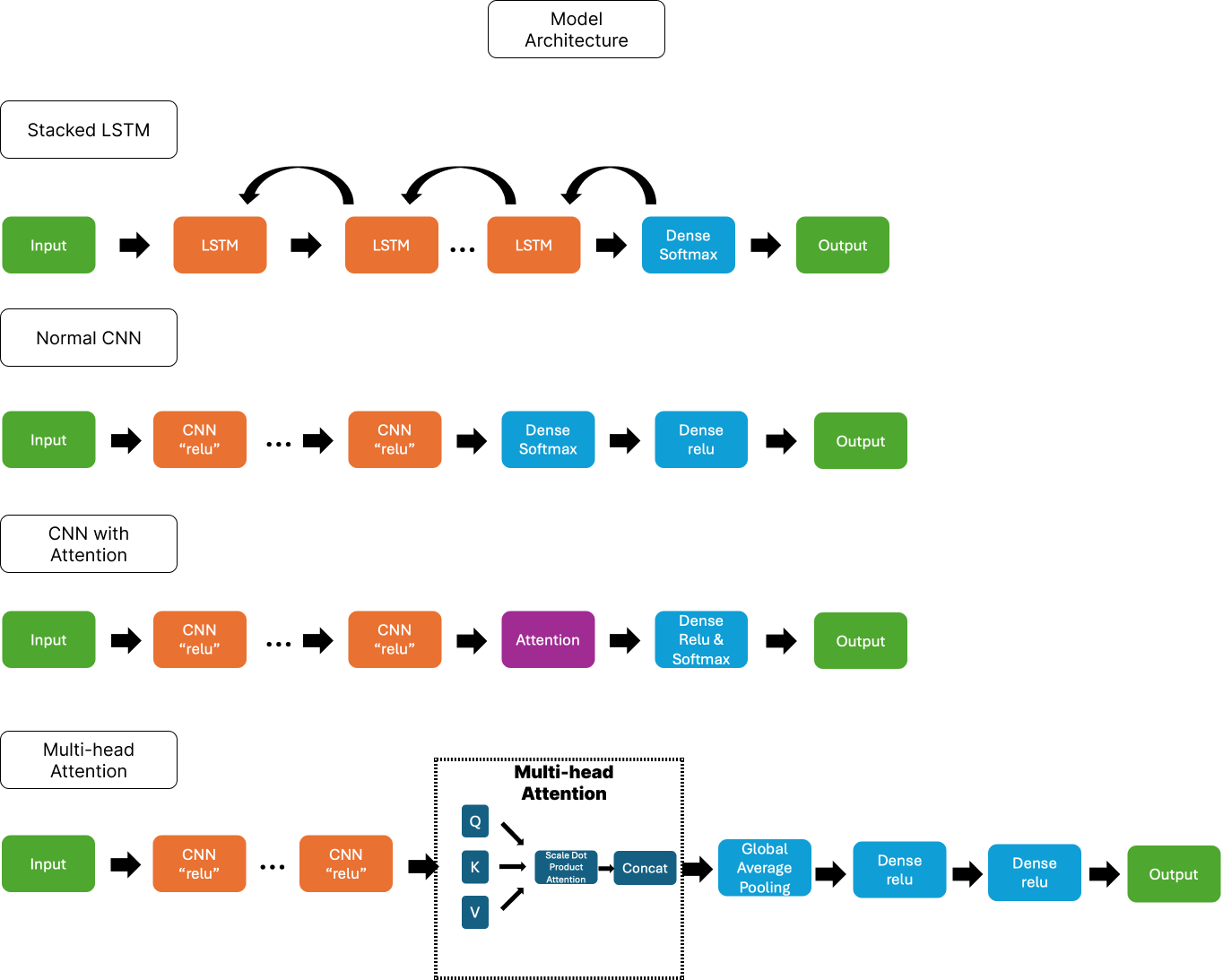}
    \caption{Three experimental model architectures}
    \label{fig:model}
\end{figure}

\begin{table}[h]
\centering
\caption{Model Performance Metrics}
\label{table:metrics}
\begin{tabular}{@{}l p{0.9cm} p{0.9cm} p{0.9cm} p{0.9cm}@{}}
\toprule
\multicolumn{5}{c}{\textbf{Sample Distribution:} Y=1 N=171, Y=0 N=2698} \\
\midrule
Model & \centering Acc. & \centering Prec. & \centering Rec. & \centering\arraybackslash F1 score \\
\midrule
LSTM (1 mod.) & \centering 0.93 & \centering 0.52 & \centering 0.50 & \centering\arraybackslash 0.49 \\
LSTM+Stat (2 mod.) & \centering 0.86 & \centering 0.62 & \centering 0.81 & \centering\arraybackslash 0.65 \\
LSTM+Stat (4 mod.) & \centering 0.86 & \centering 0.62 & \centering 0.78 & \centering\arraybackslash 0.65 \\
LSTM+Avg (4 mod.) & \centering 0.94 & \centering 0.63 & \centering 0.52 & \centering\arraybackslash 0.52 \\
\hline
CNN (1 mod.) & \centering 0.87 & \centering 0.58 & \centering 0.64 & \centering\arraybackslash 0.60 \\
CNN+Stat (2 mod.) & \centering 0.91 & \centering 0.65 & \centering 0.74 & \centering\arraybackslash 0.68 \\
CNN+Stat (4 mod.) & \centering 0.91 & \centering \textbf{0.66} & \centering \textbf{0.75} & \centering\arraybackslash 0.69 \\
CNN+Avg (4 mod.) & \centering 0.87 & \centering 0.58 & \centering 0.65 & \centering\arraybackslash 0.60 \\
\hline
CNN-Attention (1 mod.) & \centering 0.79 & \centering 0.55 & \centering 0.64 & \centering\arraybackslash 0.54 \\
CNN-Attention+Stat (2 mod.) & \centering 0.87 & \centering 0.64 & \centering 0.83 & \centering\arraybackslash 0.67 \\
CNN-Attention+Stat (4 mod.) & \centering 0.89 & \centering \textbf{0.66} & \centering \textbf{0.85} & \centering\arraybackslash 0.70 \\
CNN-Attention+Avg (4 mod.) & \centering 0.85 & \centering 0.59 & \centering 0.72 & \centering\arraybackslash 0.62 \\
\hline
Multi-head SA(1 mod.) & \centering 0.82 & \centering 0.60 & \centering 0.67 & \centering\arraybackslash 0.61 \\
Multi-head SA+Stat (2 mod.) & \centering 0.90 & \centering 0.68 & \centering 0.65 & \centering\arraybackslash 0.66 \\
Multi-head SA+Stat (4 mod.) & \centering 0.90 & \centering \textbf{0.72} & \centering \textbf{0.71} & \centering\arraybackslash 0.72 \\
Multi-head SA+Avg (4 mod.) & \centering 0.90 & \centering 0.69 & \centering 0.69 & \centering\arraybackslash 0.69 \\
\bottomrule
\end{tabular}
\end{table}

\par In our study, as delineated in Table \ref{table:metrics} and Figure \ref{fig:model}, we elucidate the efficacy of integrated strategies, ranging from foundational neural networks to an advanced BodyAttention network and CNN with Multi-head Self-Attention mechanism, on pivotal performance metrics. This investigation clearly delineates the impact of varied data fusion strategies on precision, accuracy, recall, and F1 scores during protective behavior detection. Through the integration of statistical weighting in LSTM and CNN base models, expansion to BodyAttention Network (BANet) \cite{wang2019learning}, and the adoption of Multi-head Self-Attention, we aim to evaluate the enhancement in model performance facilitated by these methodologies across diverse neural network architectures. Our research particularly explores how the amalgamation of statistical correlations with a human-centered approach can ameliorate model outcomes, especially within the challenges posed by imbalanced datasets.

\par Our findings emphasize the pivotal role of modality segmentation and decision-level fusion, informed by statistical insights, in augmenting the accuracy of protective behavior detection. They also highlight the broad applicability of statistically driven weighting strategies within various complex neural frameworks. Through an in-depth analysis of performance shifts across different configurations, we reveal the critical role of integrating sophisticated data processing techniques, human-centered perspectives, and statistical correlations. This holistic strategy markedly enhances the precision and effectiveness of protective behavior recognition in complex physiological datasets, representing a significant advance toward nuanced patient-centered healthcare solutions. Given the imbalanced nature of our dataset, we prioritize precision, recall, and F1 score over accuracy to more accurately reflect model performance across labels. We employed four distinct models— LSTM, CNN, CNN-Attention, and CNN with Multi-head Self-Attention—to diversify our evaluation and mitigate single-model reliance risks. It is important to note that a comparative performance analysis among these models is not within this paper's scope. This methodology underscores our dedication to a thorough evaluation, aiming to deepen the understanding of multi-modal data fusion's impact on model efficacy amidst dataset imbalances.


\par The data presented in Table \ref{table:metrics} provide a detailed analysis of performance differences across models using various modality integrations. A key observation is that models based on a single modality can achieve high accuracy, up to 0.93, but often at the expense of other metrics like precision, recall, and F1-score, which hover around 0.55. This indicates a potential trade-off between maximizing accuracy and ensuring balanced performance across all metrics. Examining models with dual modalities reveals different impacts on performance metrics. For example, while the LSTM model experiences a slight drop in accuracy with dual-modality integration, other architectures show improvements. This variation highlights the limitations of relying solely on a single modality, which, despite high accuracy, may not fully capture a broader range of evaluative measures. Conversely, incorporating an additional modality significantly enhances precision, recall, and F1-scores, with increases ranging from 12\% to 62\%. This strongly supports the notion that integrating multiple modalities enhances model robustness. Specifically, segregating input features into distinct spatial and sEMG modalities, rather than combining all 70 features, proves to be a strategic advantage. This approach improves the model’s ability to identify and evaluate protective behaviors, resulting in better precision and sensitivity.


\par The analysis delineates that while models based on a singular modality bypass the complexities of decision-layer weighting, rendering the choice between statistical and average weighting irrelevant, the introduction of a statistically driven weighting strategy in dual-modality configurations significantly enhances model performance by leveraging data diversity and modality-specific importance. This strategic use of statistical weighting in models integrating two modalities distinctly outperforms single-modality models, affirming the superiority of a multimodal fusion approach. This research underscores the efficacy of partitioning input features into distinct modalities coupled with the judicious use of statistical weighting at the decision layer, thereby substantially improving the precision and reliability of protective behavior detection. It paves the way for sophisticated multimodal integration in human-centered computing, setting a benchmark for handling complex, varied datasets with enhanced accuracy.

\par Adopting a human-centered approach to segment the dataset into four modalities demonstrates potential enhancements in model performance. Particularly, models employing CNN-Attention mechanisms exhibit slight but positive differences, suggesting an improved capability in capturing pain behavior features. This indicates that modality segmentation, guided by human-centered principles, can amplify model effectiveness. Further analysis reveals that models employing statistical weighting generally outperform those using mean weighting, except in the case of LSTM with four modalities employing average weighting—a scenario that mirrors the trade-offs observed in single-modality LSTM models. 

\par This investigation emphasizes the advantage of multimodal strategies, where strategic feature grouping and statistical decision-making markedly elevate model efficacy. Our findings particularly highlight the utility of attention mechanisms, like CNN-Attention and CNN with Multi-head Self-Attention, in a four-modality framework, reinforcing the benefits of human-centered modality segmentation. Overall, the transition to a four-modality model, grounded in human-centered design and statistical weighting, is validated as superior to single-modality approaches, bolstering the case for sophisticated multimodal fusion in enhancing protective behavior recognition.

\par The examination distinctly emphasizes the superiority of statistical weighting over average weighting in multimodal configurations, enhancing model performance. However, an exception was observed in the LSTM model utilizing average weighting, suggesting a nuanced balance between accuracy and other metrics. Despite this, the evidence strongly supports the advantages of the multimodal approach over singular modality frameworks. This finding reinforces the idea that strategic segmentation of modalities, when combined with statistical weighting, markedly advances protective behavior detection models.


\par This study highlights the critical importance of a human-centered modality segmentation strategy and the precise application of statistical methods in decision-making processes to optimize model outcomes. The integration of diverse modalities, guided by careful grouping and statistical weighting, not only elevates model performance but also marks a significant advance towards crafting more nuanced, interpretative models within protective behavior recognition. Consequently, this methodological approach not only pushes the boundaries of multimodal data fusion but also underlines the necessity of integrating human-centric perspectives and statistical insights into the broader narrative of human-centered computing.

\subsection{Cross Validation}

\begin{table}[h]
\centering
\caption{Leave-One-Out Cross-Validation (LOOCV)}
\label{table:loocv}
\begin{tabular}{@{}l p{0.7cm} p{0.7cm} p{0.7cm}@{}}
\toprule
Model & \centering Acc. &  \centering Rec. & \centering\arraybackslash F1 score \\
\midrule
CNN-Attention (1 mod.) & \centering 0.800 & \centering 0.631 & \centering\arraybackslash 0.547 \\
CNN-Attention+Stat (2 mod.) & \centering \textbf{0.907} & \centering \textbf{0.647} & \centering\arraybackslash \textbf{0.631} \\
CNN-Attention+Stat (4 mod.) & \centering \textbf{0.908} & \centering \textbf{0.642} & \centering\arraybackslash \textbf{0.628} \\
CNN-Attention+Avg (4 mod.) & \centering 0.886 & \centering 0.635 & \centering\arraybackslash 0.603 \\

\bottomrule
\end{tabular}
\end{table}

\par From the experimental metrics in Table \ref{table:metrics}, the CNN-Attention model demonstrates robust performance on the Emopain dataset. To substantiate the efficacy of our approach, we adopt a Leave-One-Out Cross-Validation (LOOCV) strategy, meticulously testing each dataset instance as an individual test case, with the remaining data serving for training. This technique guarantees a thorough model evaluation across diverse configurations, notably different modalities delineated by human-centered principles. By scrutinizing the CNN-Attention architecture and integrating statistical weighting for feature selection, we methodically investigate how modality fusion and statistical weighting influence model effectiveness.

\par The Leave-One-Out Cross-Validation (LOOCV) metrics in Table \ref{table:loocv} demonstrate that both dual-modality and four-modality configurations surpass the single-modality training approach, affirming the benefits of data heterogeneity segmentation and human-centered modality segmentation. Notably, the CNN-Attention model, when expanded to include two modalities with statistical weighting, shows a significant improvement in accuracy from 0.800 to 0.907 and an increase in F1-score from 0.547 to 0.631. This enhancement highlights the clear advantage of integrating multiple data sources over a single modality framework, indicating that the inclusion of diverse data types enriches the model’s performance in predicting outcomes.

\par Furthermore, transitioning from a two-modality to a four-modality configuration with statistical weighting (CNN-Attention+Stat (4 mod.)) marginally enhances the model's accuracy to 0.908, maintaining superior recall (Rec.) and F1-score metrics compared to the single-modality model. This slight improvement evidences the nuanced benefits of adopting a human-centered approach to modality segmentation, where data is divided into four distinct modalities based on its relevance and interaction with human behavioral patterns.

\par Moreover, a comparison between the four-modality configurations—statistical weighting versus average weighting (CNN-Attention+Avg (4 mod.))—reveals a distinct advantage for the former. The model employing statistical weighting (CNN-Attention+Stat (4 mod.)) achieves higher accuracy, recall, and F1-score than the model utilizing average weighting, with respective metrics of \textbf{0.908}, \textbf{0.642}, and \textbf{0.628} against 0.886, 0.635, and 0.603. This differential highlights the effectiveness of our statistical relevance weighting strategy, proving it to be a more effective method for integrating diverse modalities than merely averaging their contributions.

\par In summation, the Leave-One-Out Cross-Validation (LOOCV) findings robustly advocate for the deployment of multimodal fusion frameworks, amalgamating statistical correlations and human-centered methodologies. The analysis distinctly highlights the pivotal role of statistical weighting in augmenting model efficacy. It evidences that modality segmentation, underpinned by human-centered considerations, not only contributes positively but that the strategic application of statistical weighting across such segmented modalities markedly optimizes model performance. This approach signifies a notable progression in protective behavior detection, establishing the profound impact of integrating statistical insights with human-centered design principles on enhancing computational models within this domain.

\section{Discussion}
\par This research introduces an innovative approach to affective computing and multimodal data fusion by integrating statistical methods with human-centered computation. Our findings indicate that statistical algorithms can more effectively extract data representations across various modalities, which enhances our understanding and modeling of human behavior. Traditional machine learning methods often struggle with the complexity of human behavior and the interplay between different factors \cite{lecun2015deep}. However, our approach successfully identifies key features linked to pain behavior by combining statistical correlations with human expert knowledge. This not only enhances model performance but also improves interpretability, providing valuable insights for the development of future interactive intelligent systems \cite{holzinger2018machine}.

\par The study underscores the importance of data-driven statistical methods in pain recognition representation extraction, particularly in situations where complete reliance on patient self-reporting is not feasible. While self-reporting is a crucial means of pain recognition, it is limited by subjectivity, communication barriers, psychological factors, sociocultural influences, and feedback time delays\cite{hanson2009body}. Feature extraction based on statistical correlations provides an objective, continuous, and non-invasive method for pain representation extraction, complementing the limitations of self-reporting. This data-driven feature extraction can assist healthcare professionals in more accurately assessing patients' pain conditions, especially when patients are unable to self-report\cite{moeslund2006survey}.

\par Furthermore, given the substantial variability in human samples, personalisation has long been a potential barrier to digital healthcare. However, by utilising modality feature extraction driven by statistical correlations, our system can capture unique pain expression patterns for each individual, contributing to more precise and effective treatment strategies. This approach not only has the potential to improve patients' quality of life but may also reduce the risk of drug abuse, particularly in the management of chronic pain\cite{dworkin2005core}.

Finally, the statistical correlation-driven multimodal fusion framework and its representation learning approach that we have demonstrated possess extensive application potential, with the possibility of further extension to other domains involving complex human-centred computing. This research provides a viable framework for integrating statistical methods, machine learning, and human expertise to address complex challenges of human-computer interaction. This interdisciplinary approach not only enhances technical performance but also augments its credibility and acceptability in practical applications. As technology continues to advance, we anticipate seeing more innovative applications based on this framework, ultimately realising more intelligent and humanised interactive systems. This integrated approach opens up new possibilities for future human-machine collaboration, with the potential to play a crucial role in various complex human-centred computing tasks.

\section{Conclusion}
\par This research venture has systematically unveiled the efficacy of integrating statistical methods and human-centered perspectives within the ambit of multimodal pain behaviours recognition, employing an array of deep learning architectures including convolutional neural networks (CNN), long short-term memory networks (LSTM), CNN-Attention networks and CNN with Multi-head Self Attention. The cornerstone of our exploration was to enhance the precision and utility of complex pain recognition endeavors through the lens of statistical relevance and human-centered modality segmentation.

\par The incorporation of statistically correlated vote weights and a human-centered approach to data segmentation stands as this study's central innovation. This methodology significantly enhances model performance and paves the way for a deeper understanding and interpretation of multimodal data. It signals a shift towards choosing optimal classifiers for each modality, refining our voting strategy for the ultimate decision-making process. Considering the diversity and weak correlations among modalities, selecting a classifier tailored to the characteristics of each modality is crucial. This sophisticated strategy highlights the importance of a customized approach to modality-specific model selection, thereby boosting the efficacy of multimodal fusion.

\par Furthermore, the research integrates statistical significance with a human-centered perspective, paving the way for explainable AI in pain recognition. This approach delves into the distinct impact of each data modality on model outcomes, promoting a model of explainability that not only clarifies the workings of complex models but also enhances their adaptability and accuracy. This shift highlights our methodology's broad applicability, not only in pain management but across diverse domains of human-centered computing, spotlighting its potential to revolutionize how we interact with and leverage AI technologies.

\par In summary, this research not only pioneers the integration of statistical correlations with human-centered methods for multimodal data fusion in pain recognition but also advances the field by offering novel insights into modality fusion strategies. It enhances the discussion on improving model performance and interpretability within human-centered computing. The comprehensive framework established for multimodal fusion application spotlights the potential of statistical insights for model explainability, fostering trust in AI systems. Consequently, this work paves the way for transformative applications in AI systems closely aligned with human needs and behaviors, promising significant advancements across various human-focused domains.




%

\section*{Ethical Impact Statement}

This research intersects affective computing, multimodal data analysis, and pain recognition, aiming to enhance the precision and personalization of pain management through statistical methods. We conscientiously address the potential societal, environmental, and ethical impacts, focusing on aspects such as explainability, transparency, liability, fairness, efficacy, robustness, privacy, security, and sustainability.

\paragraph{Explainability and Transparency} Our study is committed to ensuring that advancements in pain recognition are both explainable and transparent. We leverage statistical models and human-centered perspectives to allow users and stakeholders to understand the rationale behind our predictions. This fosters trust and facilitates deeper understanding and acceptance of our technology in clinical settings.

\paragraph{Liability} 
Acknowledging the profound implications of our work in enhancing pain recognition, we recognize our responsibility to ensure the reliability and accuracy of our findings. Misinterpretations or inaccuracies could significantly impact patient care and treatment outcomes. Therefore, we adhere to stringent validation protocols, underpinned by ethical considerations, to mitigate potential risks and liabilities associated with our research outputs.

\paragraph{Fairness} Our research actively addresses the issue of fairness by incorporating the EmoPain dataset\cite{aung2015automatic} that includes chronic pain participants and healthy individuals. This diversity ensures our models do not inadvertently perpetuate biases against certain demographics. By integrating a human-centered approach, we actively seek to understand and incorporate diverse pain expressions and experiences, thereby mitigating the risk of biases that could otherwise compromise the fairness of our models.

\paragraph{Efficacy and Robustness} The intersection of statistical methods and human-centered design in our work aims to enhance the efficacy and robustness of pain recognition technologies. Our approach ensures that our models are not only accurate but also resilient to the complexities and variabilities inherent in human pain experiences, thereby supporting reliable pain management practices.

\paragraph{Privacy and Security} The sensitivity of health-related data necessitates stringent privacy and security measures. Our research select the EmoPain dataset\cite{aung2015automatic} that adheres to the highest standards of data protection, ensuring all participant data is anonymized and securely stored. Access controls, encryption, and ethical data handling practices protect against unauthorized access and data breaches.


In conclusion, our work not only signifies a step forward in the technical domain but also embodies a comprehensive ethical approach. Through our commitment to ethical considerations, we aspire to contribute meaningfully to the field of affective computing, ensuring that our innovations in pain recognition are responsible, equitable, and beneficial for all stakeholders involved.




\section*{Acknowledgment}

We’d like to thank Prof. Nadia Berthouze and Dr. Chuang Yu for helpful suggestion. This research is fully supporteds by Holistic AI and partially supported by University College London Interaction Center, University College London, UK.

\bibliographystyle{IEEEtran}
\bibliography{sample}

\end{document}